\documentclass[runningheads]{llncs}
\usepackage{graphicx}

\usepackage{textcomp}
\usepackage{times}
\usepackage[singlelinecheck=false,justification=raggedright]{caption}
\usepackage{gensymb}
\usepackage{multirow}
\usepackage{url}
\usepackage{array}
\usepackage{adjustbox}
\usepackage{threeparttable}
\usepackage{graphicx,subcaption}
\usepackage[linesnumbered,ruled,vlined]{algorithm2e}
\begin{document}
\title{MULTI-MODAL MULTI-CLASS PARKINSON DISEASE CLASSIFICATION USING CNN and DECISION LEVEL FUSION}
%
\titlerunning{MMMC PD CLASSIFICATION USING CNN and DL FUSION}
%
\author{Sushanta Kumar Sahu\inst{1} \orcidID{0000-0002-4384-8939} \and \\
Ananda S. Chowdhury\inst{1}\orcidID{0000-0002-5799-3467}} 

\author{Sushanta Kumar Sahu \orcidID{0000-0002-4384-8939} \and \\
Ananda S. Chowdhury\orcidID{0000-0002-5799-3467}}

\authorrunning{S. Sahu et al.}
\institute{Jadavpur University, Kolkata, West Bengal, India 
\\ \email{\{sksahu.etce.rs, as.chowdhury\}@jadavpuruniversity.in}
}

%
\maketitle              
%
\begin{abstract}
Parkinson’s disease (PD) is the second most common neurodegenerative disorder, as reported by the World Health Organization (WHO).
In this paper, we propose a direct three-Class PD classification using two different modalities, namely,  MRI and DTI. The three classes used for classification are PD, Scans Without Evidence of Dopamine Deficit (SWEDD) and Healthy Control (HC). We use white matter (WM) and gray matter (GM) from the MRI and fractional anisotropy (FA) and mean diffusivity (MD) from the DTI to achieve our goal. We train four separate CNNs on the above four types of data. At the decision level, the outputs of the four CNN models are fused with an optimal weighted average fusion technique. We achieve an accuracy of 95.53\% for the direct three-class classification of PD, HC and SWEDD on the publicly available PPMI database. Extensive comparisons including a series of ablation studies clearly demonstrate the effectiveness of our proposed solution.
\end{abstract}
\vspace*{-0.7cm}
\section{Introduction}
\label{section:intro}
Parkinson's disease is the second most common neurological disorder that affects movement and can cause tremors, stiffness, and difficulty with coordination ~\cite{Adeli2016}. Early diagnosis of PD is important for effective treatment, as there is currently no cure for the disease. However, diagnosis can be challenging due to the variability of symptoms and lack of definitive biomarkers. According to the World Health Organization (WHO), PD affects approximately 1\% of people aged 60 years and older worldwide. 
However there are approximately 10\% of clinically diagnosed patients with early stage PD who exhibit normal dopaminergic functional scans. This class, which signifies a medical condition distinct from PD, is known as Scans Without Evidence of Dopamine Deficit (SWEDD) \cite{singh2018,kim2016using}. As a result of the evolution of this new class, difficulty of diagnosing PD has increased manifold, leading to a three-class classification problem of PD vs. SWEDD vs. HC with class overlaps~\cite{kim2016using}.

MRI, SPECT and PET are commonly used imaging techniques for PD diagnosis. However, PET and SPECT are not preferred by doctors due to invasiveness and cost ~\cite{Long2012}. DTI is a newer technique that measures water molecule movement to analyze white matter microstructure which gets affected in PD. In the literature, quite a few works were reported on PD classification based on machine learning (ML) and deep learning (DL) models applied to neuroimaging data. Salat et al. \cite{salat2009age} found correlations between gray or white matter changes and age using Voxel-based Morphometry (VBM). Adeli et al.~\cite{Adeli2016} used a recursive feature elimination approach for two-class classification with 81.9 \% accuracy. Cigdem et al.~\cite{Cigdem2018} proposed a total intracranial volume method with 93.7\% accuracy. Singh et al.~\cite{singh2018} presented a ML framework for three two-class classifications.  Chakraborty et al.~\cite{chakraborty2020} presented an DL model with 95.29\% accuracy. A DL-based ensemble learning technique was reported by~\cite{rajanbabu2022} with 97.8\% accuracy.

Recent research indicates that combining features from more than one imaging modality can improve the classification accuracy. For example, Li et al.~\cite{li2014discriminative} showed that combining DTI and MRI features improves the classification accuracy in Alzheimer's disease. The authors of \cite{li2019} used MRI and DTI, but only considered MD data from DTI for PD classification. They used a stacked sparse auto-encoder to achieve better classification accuracy. In light of the above findings, we anticipate that MRI and DTI can be effectively combined to better analyze PD. To increase decision accuracy, we mention here a few decision level fusion techniques. 
Majority voting technique is the most common techniques used in late fusion ~\cite{daskalakis2009}. This strategy, however, may not be appropriate for multi-class classification applications. Single classifiers work well on most subjects, but error rates are enhanced for some difficult-to-classify subjects due to overlap across many categories. Instead of using a majority voting strategy, a modified scheme called modulated rank averaging is employed in~\cite{DE2021114338}. We feel the accuracy may be enhanced even further by fine-tuning the weights computed in the modulated rank averaging approach.

As a summary, we can say that there is a clear dearth of direct three-class PD classification strategies and that too with multi-modal data. In this paper, we present a direct three-class PD classification using CNNs and decision level fusion. We investigate full potential of multimodal data \textit{i.e.,} FA \& MD from DTI and WM \& GM from MRI. We use four CNNs to analyze these four types of data. Outputs from all these four models are finally fused using an Optimal Weighted Average Fusion (OWAF) technique at the decision level. Since neuroimaging datasets are small, data augmentation is adopted to ensure proper training with the CNNS~\cite{DE2021114338}. We now summarize our contributions as below:
\begin{enumerate}
\item We address a direct three-class classification task (PD, HC and SWEDD) for Parkinson's disease, which is certainly more challenging than the current trend of a single binary classification (for 2-class problem) or multiple binary classifications (for 3-class problem).
\item We make effective use of the underlying potential of multi-modal neuroimaging, namely T1-weighted MRI and DTI. In particular, we train four different CNNs on WM, GM data from MRI and FA, MD data from DTI. Such in-depth analysis of multi-modal neuroimaging data is largely missing in the analysis of PD.
\item Finally, at the decision level, the outputs of each CNN model are fused using an Optimal Weighted Average Fusion (OWAF) strategy to achieve state-of-the-art classification accuracy.
\end{enumerate}
\vspace*{-0.7cm}
\section{Proposed method}
\label{section:Proposed method}
\begin{figure}
\centering
\captionsetup{justification=centering}
\includegraphics[width=0.8\textwidth, height=4.5cm]{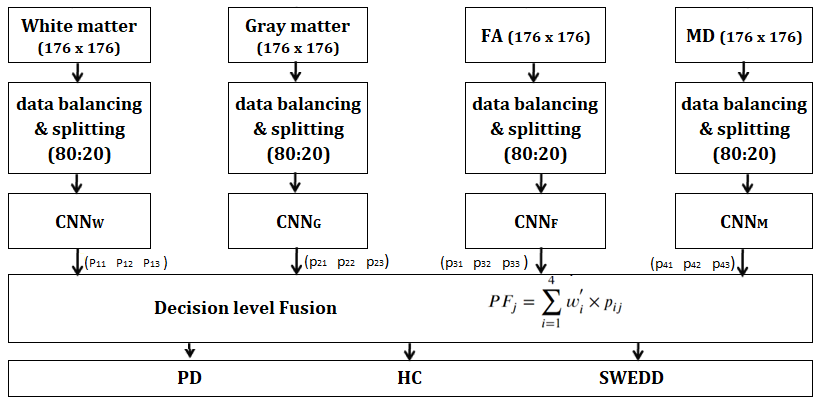}
\caption{Direct three-class Parkinson’s disease classification framework}
\label{fig:solution_pipeline}
\end{figure}

 Our solution pipeline for an end-to-end direct three-class classification of PD from DTI and MRI consists of four CNN networks. Each CNN network yields a $3\times1$ probability vector, which represents the probability that the data falls into one of three classes, \textit{i.e.}, PD, HC or SWEDD. The probability vectors are then combined using the OWAF technique. In section \ref{subsection:Data_Pre_Proc}, we discuss how WM and GM are obtained from MRI data and MD and FA are used from DTI data. Section \ref{subsection:Data_balancing} describes ADASYN, an oversampling strategy. Section \ref{subsection:CNN_Architecture} presents the proposed CNN architecture. In section \ref{subsection:Fusion of CNN network}, we discuss the decision level fusion. Figure \ref{fig:solution_pipeline} illustrates the overall pipeline of our solution.
 \vspace*{-0.5cm}
\vspace*{-1cm}
\subsection{Data Pre-processing}
\label{subsection:Data_Pre_Proc}
In this work, voxel-based morphometry (VBM) is used to prepare MRI data. The data is preprocessed using SPM-12 tools and images are normalized using the diffeomorphic anatomical registration with exponentiated lie algebra (DARTEL) method ~\cite{ashburner2007}. This SPM-12 tool segments the whole MRI data into GM, WM and cerebrospinal fluid, as well as the anatomical normalization of all images to the same stereotactic space employing linear affine translation, non-linear warping, smoothing and statistical analysis. After registration, GM and WM volumetric images were obtained and the unmodulated image is defined as the density map of grey matter (GMD) and white matter (WMD). The PPMI database contains all information regarding DTI indices, including FA and MD. Brain scans from PD groups as well as HC have distinct voxel MD values. MD and FA can be expressed mathematically expressed as  ~\cite{alexander2007diffusion}: 
\begin{equation}
    \label{eqMD}
    MD=\frac{\lambda_1+\lambda_2 +\lambda_3}{3}=\frac{D_{xx}+D_{yy}+D_{zz}}{3}
\end{equation}
\begin{equation}
    \label{eqFA}
    FA=\sqrt{\frac{1}{2}}\sqrt{\frac{{(\lambda_1-\lambda_2)}^2+{(\lambda_2-\lambda_3)}^2+{(\lambda_3-\lambda_1)}^2}{{\lambda_1}^2+{\lambda_2}^2+{\lambda_3}^2}}
\end{equation}
In equations \ref{eqMD} the diagonal terms of the diffusion tensor are $D_{xx}, D_{yy} and D_{zz}$ and the sum of these diagonal terms constitutes its trace. After prepossessing, four types of data are made available, namely, grey matter (GM), white matter (WM), fractional anisotropy (FA) and mean diffusivity (MD).
\vspace*{-0.5cm}
\begin{figure}
\includegraphics[width=\textwidth]{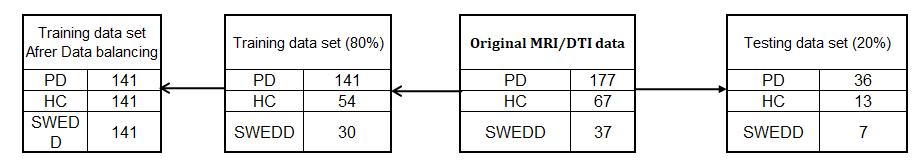}
\caption{Dataset division strategy}
\label{fig:data_division}
\end{figure}
\vspace*{-1.65cm}
\subsection{Data balancing with ADASYN}
\label{subsection:Data_balancing}
We find each of GM, WM, FA, MD to be highly imbalanced across the three classes. Further, the number of training samples required to feed a DL model is insufficient. So, ADASYN, an oversampling method is used to increase the number of samples for each minority class ~\cite{pristyanto2022}. The primary idea behind the use of an ADASYN technique is to compute the weighted distribution of minority samples based on a wide range of out-of-elegance neighbors. 
A difficult-to-train minority instance surrounded by more out-of-class examples is given a better chance of being augmented through producing synthetic samples. Using a set of pseudo-probabilistic rules, a predetermined number of instances are generated for every minority class depending on the weighted distribution of its neighbors. Following the implementation of this up-sampling approach, the total number of samples in each of the classes is 141 volumetric images. The details of data set division strategies are shown in Figure \ref{fig:data_division}.

\indent Let a dataset consists of $m$ samples of the form $({x_i,y_i})$; where $i \in [1,m]$, $x_i$ being the $i^{th}$ sample of the n-dimensional feature space X and $y_i$ being the label (class) of the sample ${x_i}$. Let $m_{min}$ and $m_{maj}$ be number of the minority and majority class samples respectively. Then the ratio of minority to majority sample will be expressed as:  d = $m_{min}$ /$m_{maj}$. Let, $\beta$ be the balance level of the synthetic samples. So, $\beta = 1$ represents that both the classes are balanced using ADASYN. Number of synthetic minority data to be created is given by: \ref{adasyn1}.
\begin{equation}
    \label{adasyn1}
  G=(m_{maj} - m_{min}) \times \beta
\end{equation}
For $\beta \neq 1$, synthetic samples will be created for each set of minority data based on the Euclidean distance of their k-nearest neighbors. The dominates of the majority class in each neighborhood is expressed as: $r_i= \frac {\Delta_i} k$. where $\Delta_i$ is the number of examples in the k-nearest neighbors of $x_i$ that belong to the majority class. Higher value of ${r_i}$ in a neighborhood indicates more examples of the majority class which makes them harder to learn. We next determine how many synthetic samples per neighborhood need to be generated as: $G_i= G \times \hat{r_i}$. Here $\hat{r_i}$ represent normalized version of ${r_i}$. This captures the adaptive nature of the ADASYN algorithm, which means more data is created for harder to learn neighbourhoods. We generate $G_i$ number of synthetic data ${s_{i,j}, j=1,2, \cdots, G_i}$ for each neighbourhood $i$ as shown in equation \ref{adasyn4} below:
\begin{equation}
    \label{adasyn4}
   {s_{i,j}}= x_i + (x_{zi}- x_i ) \times \lambda
\end{equation}
Where $x_i$ and $x_{zi}$ are two minority occurrences in the same neighborhood and $\lambda$ is a random integer between 0 and 1. In our work, these synthetic images are generated with the help of the Scikit-Learn library such that all the classes are balanced in nature.
\vspace*{-0.7cm}
\subsection{Proposed CNN Architecture}
\label{subsection:CNN_Architecture}
\begin{figure}
\centering
\includegraphics[width=0.9\textwidth, height=3.6cm]{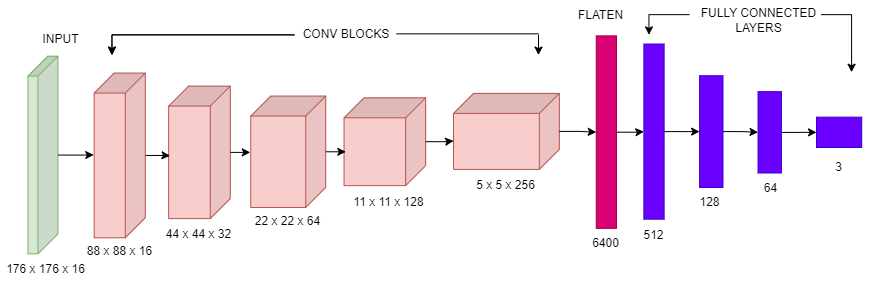}
\caption{Architecture of the proposed CNN}
\label{fig:cnn}
\end{figure}

We use four CNN models, each with ten convolutional layers and four dense (FC) layers, for the direct three-class classification task of PD. The proposed network's architecture is depicted in Figure \ref{fig:cnn}. We chose fewer parameters in the proposed architecture than in the original VGG16 ~\cite{simonyan2014very} by decreasing network depth. Our proposed network is similar to that proposed in ~\cite{advian2021}. The number of layers are limited to maintain a trade-off between accuracy and computational cost. To generate feature representations of brain MR scans, convolution layers are used. The final FC layer and a soft max operation are used for the classification task. The volumetric input data is processed slice-wise, with each slice having a size of $176 \times 176$ pixels. 
We have used max-pooling in the pooling layers to reduce the image size. The flattened layers convert the reduced feature maps to a one-dimensional feature map. The fully connected layers  classify this feature map into three classes: HC, PD and SWEDD. The cross-entropy loss function (CELF) is the most common loss function used for classification problems since it has better convergence speeds for training deep CNNs than MSE and hinge loss.
As a result, we consider CELF for this work, which is mathematically expressed as:
\begin{equation}
\label{eq7}
L_{CELF} = - \sum\limits_{i = 1}^{N}\sum\limits_{j = 1}^{K}p(y_{j}|\mathbf{x})\log\widehat{p}\left( {y_{j}|\mathbf{x}} \right)
\end{equation}
where $p(y_{j}|\mathbf{x})$ is the original class label distribution and $\widehat{p}(y_{j}|\mathbf{x})$ is the predicted label distribution from the CNN network. Here $N$ and $K$ represent the number of models and number of classes respectively. For our problem, $N=4$ and $K=3$.
\vspace*{-0.5cm}
\vspace*{-1cm}
\subsection{Decision Level Fusion of CNN Networks}
\label{subsection:Fusion of CNN network}
We use four CNN models, one each for GM, WM, FA and MD.
We fuse these predicted probabilities with the help of suitable weights. The weights are generated in two stages. In the first stage, the weights are generated using the modulated rank averaging (MRA) method ~\cite{DE2021114338}. The weights in the MRA method are given by:
\begin{equation}
\label{eq_mra}
w_i=\frac{f_i}{\sum_{i=1}^{N-1}f_i + R_{max}}
\end{equation}
\noindent In equation \ref{eq_mra}, $f_i$ and $R_{max}$ indicate the normalizing factor and the rank of the model having highest accuracy respectively. The normalizing factor is calculated based on the rank of the current model and the difference between the accuracy of the current and next model. In the second stage, these weights are optimised using the grid search method ~\cite{FAYED2019}. Let the final optimized weights be denoted by $w_{i}^{'}$. Note that this weight vector is fixed for all 3 classes.
We combine this optimal weight vector with the respective probability vectors to obtain the overall probability of occurrence of respective classes. Let us denote by  $PF_j; j= 1,2,3$, the overall probability of occurrence of the $j^{th}$ class as a result of fusion.  
\begin{equation}
    \label{eq9}
    PF_j = \sum_{i=1}^{4} w_{i}^{'}\times p_{ij}
   \end{equation}
The final class will be the one for which $PF_j$ is maximum.
\vspace*{-0.9cm}
\begin{table}
\centering
\caption{Demographics of the subjects}
\label{tab:Image Acquisition}
\begin{tabular}{lccc}
\cline{2-4}
\multicolumn{1}{l|}{} & \multicolumn{1}{l|}{Healthy Control} & \multicolumn{1}{c|}{Parkinson's Disease} & \multicolumn{1}{c|}{SWEDD} \\ \hline
\multicolumn{1}{|l|}{Number of subjects} & \multicolumn{1}{c|}{67} & \multicolumn{1}{c|}{177} & \multicolumn{1}{c|}{37} \\ \hline
\multicolumn{1}{|l|}{Sex (Female/Male)} & \multicolumn{1}{c|}{24/43} & \multicolumn{1}{c|}{65/122} & \multicolumn{1}{c|}{14/23} \\ \hline
\multicolumn{1}{|l|}{Age*} & \multicolumn{1}{c|}{60.12 ±10.71} & \multicolumn{1}{c|}{61.24 ±9.47} & \multicolumn{1}{c|}{59.97 ±10.71} \\ \hline
\multicolumn{4}{l}{*mean ± standard   deviation}
\end{tabular}
\end{table}
\vspace*{-1.2 cm}
\section{Experimental Results}
\label{section:Experimental_Result}
In this section we present the experimental results for three class PD classification. The section is divided into three subsections. In the first subsection, we provide an overview of the PPMI database. We then discuss data preparation, computing configuration, parameter settings of ADASYN and CNN and evaluation metrics. The second subsection shows a series of ablation studies to demonstrate separate impacts of both MRI and DTI data and the proposed fusion strategy. Finally, we include comparisons with several state-of-the-art methods to demonstrate the effectiveness of the proposed solution.
\vspace*{-0.7cm}
\subsection{Data preparation and implementation details}
\label{section:data_preparation}
In our study, we included 281 subjects with baseline visits having both DTI and MRI data from PPMI. This includes 67 HC, 177 PD and 37 SWEDD subjects. Table \ref{tab:Image Acquisition} shows the demographics of the individuals used in this investigation. For all experiments, we use a system with 16.0 GB DDR4 RAM, an Intel® Core™  i7-10750H CPU @ 2.60GHz and GPU of NVIDIA GeForse RTX 3060 @ 6GB. Here, 80\% of the data were randomly selected from the PD, SWEDD and HC groups to produce the training set from 225 volumetric image scans. The remaining 56 volumetric image scans, representing approximately 20\% of each class, were utilized to create the test set. For ADASYN, we experiment with different neighbor counts (k) on the training set while keeping other settings unaltered. We find that k = 30 produced the best results. Figure \ref{fig:data_division} shows the details of data set division strategies. As a result of this, the total number of volumetric images becomes 423, where each of the three classes have 141 volumetric images. For the four CNN models, one each on GM, WM, FA and MD data, we train the network for 100 epochs. ADAM optimizer and ReLU activation functions are employed. The learning rate is initialized at $1\times10^{-4}$ with a batch size of 32.
We use the same training parameters for each model (WM, GM, FA and MD) such that there are no conflicts when we combine the outputs of the models and make a fusion at the decision level. 
We evaluate the classification performance using four different metrics. These are accuracy, precision, recall and F1 score. 
All the measures are calculated using the Scikit Learn packages ~\cite{scikit-learn}.
\vspace*{-0.80cm}
\begin{table}[htpb]
\centering
\caption{Impact of DTI and MRI data on direct 3-class classification}
\label{tab:table_2}
\resizebox{0.9\columnwidth}{!}{
\begin{tabular}{|l|c|c|c|c|}
\hline
\multicolumn{1}{|c|}{Data} &
  Accuracy in (\%) &
  Precision in (\%) &
  Recall in (\%) &
  {F1 Score in (\%) } \\ \hline
MRI (WM)  & 88.6  & 86.57 & 82.50 & 88.94 \\ \hline
MRI (GM)   & 84.2  & 81.76 & 82.50 & 84.80 \\ \hline
DTI (MD)     & 88.2  & 88.85 & 89.50 & 88.11 \\ \hline
DTI(FA) & 80.94 & 80.35 & 81.00 & 81.59 \\ \hline
WM and GM (MRI) & 90.02 & 90.17 & 90.02 & 90.09 \\ \hline
FA and MD  (DTI) & 91.14  & 91.23 & 91.14 & 91.18 \\ \hline
\begin{tabular}[c]{@{}l@{}}WM, GM, FA,  MD \\(MRI + DTI) using OWAF\end{tabular} &
  \textbf{95.53} &
  \textbf{93.64} &
  \textbf{91.99} &
  \textbf{92.74} \\ \hline
\end{tabular}}
\end{table}
\vspace*{-1.5cm}
\begin{table}[htpb]
\renewcommand{\arraystretch}{1}
\centering
\caption{Impact of fusion strategies on direct 3-class classification}
\label{tab:table_4}
\begin{tabular}{|l|c|c|c|c|}
\hline
\multicolumn{1}{|c|}{Fusion Technique} &
  \begin{tabular}[c]{@{}c@{}}Accuracy in (\%)\end{tabular} &
  \begin{tabular}[c]{@{}c@{}}Precision in (\%)\end{tabular} &
  \begin{tabular}[c]{@{}c@{}}Recall in (\%)\end{tabular} &
  \begin{tabular}[c]{@{}c@{}}F1 Measure in (\%)\end{tabular} \\ \hline
Majority Voting   & 92.19 & 92.5 & \textbf{94.87} & 92.34  \\ \hline
Model Average Fusion     & 88.6  & 82.5 & 86.57 & 85.44   \\ \hline
Modulated Rank Average   & 94.6  & 91.5 & 93.92 & \textbf{93.02}    \\ \hline
\begin{tabular}[c]{@{}l@{}} OWAF (proposed)\end{tabular} & \textbf{95.53} &  \textbf{93.64} &  91.99 &  92.74   \\ \hline
\end{tabular}
\end{table}
\vspace*{-1cm}
\subsection{Ablation Studies}
We include two ablation studies. The first study demonstrates the utility of using both MRI and DTI data. The second study conveys the benefit of OWAF, the proposed fusion strategy. 
The four CNNs are trained and evaluated on both single and multi-modal data from MRI and DTI. Our goal is to investigate the effects of using single and multi-modal data on direct three-class. The results of direct three-class classification are presented in Table \ref{tab:table_2}. This tables clearly illustrate that use of both DTI and MRI yields superior results as compared to using MRI and DTI in isolation. Note that the magnitude of improvement from the use of multi-modal data is clearly more significant for the more challenging three-class classification problem.
The four CNNs are combined using four different fusion strategies at the decision level, namely; majority voting, model average fusion, modulated rank averaging ~\cite{DE2021114338} and the proposed optimal weighted average fusion (OWAF) based on the grid search approach. When voting techniques are applied, the universal decision rule is established simply by fusing the decisions made by separate models.
In the model average fusion method, the output probabilities of each model are simply multiplied by the weight provided to that model based on its accuracy. In the modulated rank averaging method (MRA), the output probabilities of each model are updated using a weight generated based on their rank and the difference in probabilities between the models. This method gives better results than the model average fusion method and the majority voting method. In this work, we take the weights generated using MRA as the initial weights. These base weights are further optimised using the grid search method. In grid search, we fine tune the weights ($w_i$) in the range of $w_i\pm 0.05$ with a step size of 0.01. The final weights are used in our OWAF technique.
Table \ref{tab:table_4} illustrates the effects of various fusion strategies. For fare comparison, we use both MRI and DTI data in all cases. The experimental results in the table \ref{tab:table_4} clearly reveal that the proposed OWAF outperforms other fusion strategies.
\vspace*{-0.7cm}
\begin{table}[]
\renewcommand{\arraystretch}{1.4}
\centering
\caption{Comparisons of the proposed method with State-of-the-art Approaches}
\label{tab:state_of_art_compare}
\resizebox{\columnwidth}{!}{%
\begin{tabular}{|c|c|c|ccc|c|c|c|}
\hline
\multirow{2}{*}{Approach} &
  \multirow{2}{*}{ML/DL} &
  \multirow{2}{*}{MODALITY} &
  \multicolumn{3}{c|}{PD vs   HC} &
  PD vs. SWEDD &
  HC vs. SWEDD &
  PD vs. HC vs   SWEDD \\ \cline{4-9} 
                      &           &           & \multicolumn{1}{c|}{Ac}    & \multicolumn{1}{c|}{Pr}   & Re   & Ac    & Ac    & Ac \\ \hline
Adeli 2016 \cite{Adeli2016}        & ML        & MRI       & \multicolumn{1}{c|}{81.9}  & \multicolumn{1}{c|}{-}    & -    & -     & -     & -  \\ \hline
Cigdem 2018 \cite{Cigdem2018}       & ML        & MRI       & \multicolumn{1}{c|}{93.7}  & \multicolumn{1}{c|}{-}    & 95   & -     & -     & -  \\ \hline
Prashanth 2018 \cite{prashanth2018}   & ML        & SPECT     & \multicolumn{1}{c|}{95}    & \multicolumn{1}{c|}{-}    & 96.7 & -     & -     & -  \\ \hline
Singh 2018 \cite{singh2018}         & ML        & MRI       & \multicolumn{1}{c|}{95.37} & \multicolumn{1}{c|}{-}    & -    & 96.04 & 93.03 & -  \\ \hline
Gabriel 2021 \cite{Gabriel2021} &
  ML &
  MRI &
  \multicolumn{1}{c|}{\begin{tabular}[c]{@{}c@{}}99.01(M)\\      87.10(F)\\      93.05 (A)\end{tabular}} &
  \multicolumn{1}{c|}{\begin{tabular}[c]{@{}c@{}}100(M)\\      97.2(F)\\      -\end{tabular}} &
  \begin{tabular}[c]{@{}c@{}}99.3(M)\\      100(F)\\      -\end{tabular} &
  - &
  - &
  - \\ \hline
Li 2019 \cite{li2019}          & DL   (AE) & MRI + DTI & \multicolumn{1}{c|}{85.24} & \multicolumn{1}{c|}{95.8} & 68.1 & -     & 89.67 & -  \\ \hline
Tremblay 2020 \cite{tremblay2020}    & DL   & MRI       & \multicolumn{1}{c|}{88.3}  & \multicolumn{1}{c|}{88.2} & 88.4 & -     & -     & -  \\ \hline
Chakraborty 2020 \cite{chakraborty2020}  & DL   & MRI       & \multicolumn{1}{c|}{95.3}  & \multicolumn{1}{c|}{92.7} & 91.4 & -     & -     & -  \\ \hline
Sivaranjini 2020 \cite{sivaranjini2020} & DL   & MRI       & \multicolumn{1}{c|}{88.9}  & \multicolumn{1}{c|}{-}    & 89.3 & -     & -     & -  \\ \hline
Rajanbabu 2022 \cite{rajanbabu2022}    & DL   (EL) & MRI       & \multicolumn{1}{c|}{97.5}  & \multicolumn{1}{c|}{97.9} & 97.1 & -     & -     & -  \\ \hline
Proposed method &
  DL  &
  MRI + DTI &
  \multicolumn{1}{c|}{97.8} &
  \multicolumn{1}{c|}{97.2} &
  97.6 &
  94.5 &
  95.7 &
  \begin{tabular}[c]{@{}c@{}}95.53\\      Pr: 93.64\\      Re: 91.99\end{tabular} \\ \hline
\end{tabular}%
}
\begin{tablenotes}
      \item{Ac, Pr, Re, M, F, A, AE, EL, -} Indicates Accuracy, Precision, Recall, Male, Female, Average, Auto Encoder, Ensemble Learning \& \textit{data} not available respectively. All the values are in \%.
\end{tablenotes}
\end{table}
\vspace*{-1.0 cm}
\subsection{Comparisons with State-of-the-art Approaches}
We compare our method with ten state-of-the-art approaches. There are no results available for a direct 3-class PD classification. So, we compare our results with those papers that have addressed the PD classification on the PPMI database using single or multiple modalities and with three or fewer two-class classifications. The results of comparisons are shown in Table \ref{tab:state_of_art_compare}. Out of the ten methods we have considered, five are based on  machine learning (ML) and the rest five are based on deep learning (DL). Further, in four out of five DL based approaches, only a single modality, namely, MRI is used for classification. Also note that eight of these ten techniques have only addressed a single two-class classification problem between PD and HC and did not consider the challenging SWEDD class at all. The remaining two approaches did consider SWEDD as a third class but have divided the three-class classification problem into multiple binary classes \cite{singh2018,li2019}. However, Li at al.~\cite{li2019} did not report the classification results for PD vs. SWEDD in their paper. In order to have fair comparisons, we have also included three binary classifications as obtained from our method in this table. 
Our direct three-class classification accuracy turns out to be superior than two-class classification accuracy of at least eight out of 10 methods. It is also higher than two out of three binary classification accuracy of~\cite{singh2018}. Note that in~\cite{singh2018}, the authors used a somewhat different experimental protocol by considering two publicly available databases of ADNI and PPMI. In our work, we explicitly consider data with both MRI and DTI for the same individual as available solely in the PPMI database. Though the authors in~\cite{Gabriel2021} reported superior classification accuracy for male, the accuracy is much less for female and also for the average case (both male and female taken into account). Only, \cite{rajanbabu2022} has reported a higher classification accuracy than ours. But, they have considered only a single binary class classification of PD vs HC and ignored the more challenging SWEDD class. If we consider the binary classification results of our method, then we straightway outperform nine of the ten state-of-the-art competitors and even beat the remaining method~\cite{singh2018} in two out of three classifications.
\section{Conclusion}
\label{section:conclusion}
In this paper, we presented an automated solution for the direct three-class classification of PD using both MRI and DTI data. Four different CNNs were used for separate  classifications with WM and GM data from MRI and MD and FA data from DTI. An optimal weighted average decision fusion method was applied next to integrate the individual classification outcomes. An overall three-class classification accuracy of 95.53\% is achieved. Extensive testing, including a number of ablation studies on the publicly available PPMI database clearly establishes the efficacy of our proposed formulation.
 
In future, we plan to deploy our model in clinical practice with data from other neuro-imaging modalities. We also plan to to extend our model for classifying other Parkinsonian syndromes.
\bibliographystyle{elsarticle-num}
\bibliography{PReMI}
\end{document}